\documentclass[11pt]{article}
\IfFileExists{latex/acl.sty}{\def\paperpath{latex/}}{\def\paperpath{}}
\let\gambitbibliographystyle\bibliographystyle
\renewcommand{\bibliographystyle}[1]{\gambitbibliographystyle{\paperpath#1}}
\IfFileExists{latex/acl.sty}{\usepackage[final]{latex/acl}}{\usepackage[final]{acl}}
\usepackage{iftex}
\ifPDFTeX
  \usepackage[T1]{fontenc}\usepackage[utf8]{inputenc}\usepackage{times}
\else
  \usepackage{fontspec}
  \IfFontExistsTF{TeX Gyre Termes}{\setmainfont{TeX Gyre Termes}}{\setmainfont{Times New Roman}}
\fi
\usepackage{latexsym,amsmath,amssymb,booktabs,tabularx,array,graphicx,microtype}
\usepackage{tikz}
\usepackage{url}
\newcolumntype{Y}{>{\raggedright\arraybackslash}X}
\definecolor{signedblue}{RGB}{43,86,130}
\definecolor{absorange}{RGB}{190,112,28}
\title{Benchmarking Gender Bias in Machine Translation Evaluation Metrics\\across Occupations}
\author{
  Orfeas Menis Mastromichalakis\textsuperscript{1} \thanks{Corresponding author: \href{mailto:menisorfeas@gmail.com}{menisorfeas@gmail.com}.} \quad
  Giorgos Filandrianos\textsuperscript{2} \quad
  Wafaa Mohammed\textsuperscript{3} \\
  \textbf{Giuseppe Attanasio\textsuperscript{1}} \quad
  \textbf{Chrysoula Zerva\textsuperscript{2}} \\
  \textsuperscript{1}Instituto de Telecomunicações, Lisbon, Portugal \\
  \textsuperscript{2}National Technical University of Athens, Greece \\
  \textsuperscript{3}University of Amsterdam, Netherlands \\
}

\begin{document}
\maketitle
\begin{abstract}
Gender bias remains a persistent concern in machine translation (MT), affecting both generated translations and their automatic evaluation. When a source text leaves a person's gender unspecified, translations may realize that person using masculine or feminine forms, and both MT systems and evaluation metrics may exhibit systematic preferences between these alternatives despite the source providing no basis for such a distinction. We study this behavior in the WMT 2026 Automated Translation Quality Evaluation Systems Shared Task using an occupation-balanced subset of GAMBIT+\footnote{The dataset is available at: \url{https://huggingface.co/datasets/ailsntua/gambit-plus}.}. We consider seven English-source language pairs, six from the original dataset, targeting Arabic, Czech, Greek, Icelandic, Russian, and Ukrainian, and extend the original resource with German. The subset contains 1,308 masculine/feminine translation pairs per target language, with three examples for each of the 436 ISCO-08 occupational groups. We evaluate shared-task submissions and baselines for score prediction and error annotation, examining the direction, magnitude, and frequency of gender-related differences. We find an overall tendency for masculine translations to receive higher scores, as well as differences per occupation following stereotypical gender representations, although the strength and consistency of this preference vary considerably across evaluators and languages. Our results show that gender bias remains present in MT evaluation, but that capturing its extent requires looking beyond a single aggregate measure to complementary dimensions of evaluator behavior.
\end{abstract}

\section{Introduction}

Gender-related preferences can arise both in machine translation outputs and in the metrics used to assess their quality. When a source text leaves a person's gender unspecified, masculine and feminine translations may both be valid, yet MT systems may systematically favor one realization over the other \citep{menis2025assumedpublished}. This is common for gender-ambiguous occupational terms in English, such as \textit{doctor}, \textit{teacher}, or \textit{legislator}, where the source itself provides no evidence for assigning either gender. Gender-neutral or gender-fair formulations may also be possible in many cases; however, the present work focuses specifically on the contrast between masculine and feminine realizations.

The same concern extends to automatic evaluation. If two translations preserve the information expressed in the source and differ primarily in the gender used to realize an otherwise ambiguous case, an evaluation metric should not systematically reward one variant over the other. Previous work has nevertheless shown that MT evaluation metrics can exhibit such preferences \citep{filandrianos2025gambitplus, zaranis-etal-2025-watching}. This is particularly relevant because automatic metrics are widely used to compare MT systems, rank candidate translations, and guide model development, meaning that systematic evaluator preferences can affect how MT quality is ultimately progressing.

GAMBIT+ \citep{filandrianos2025gambitplus} was introduced to study this form of evaluator bias through controlled masculine/feminine translation pairs. Building on GAMBIT \citep{menis2025assumedpublished}, it organizes gender-ambiguous occupational references according to the 436 four-digit groups of the ISCO-08 classification\footnote{https://ilostat.ilo.org/methods/concepts-and-definitions/classification-occupation/} and provides paired translations in which the relevant occupation is realized once in masculine and once in feminine form. The original study evaluated 33 source--target language pairs, combining three source languages with eleven target languages, and used the resulting challenge set to analyze quality estimation metrics at WMT 2025.

In this work, we revisit this evaluation for the WMT 2026 Automated Translation Quality Evaluation Systems Shared Task \cite{wmt26-qemetrics} using a smaller, occupation-balanced subset of GAMBIT+. The full benchmark contains nearly 290,000 paired instances, making large-scale evaluation increasingly costly as the number and complexity of evaluation systems grows. We therefore retain three examples for each of the 436 ISCO-08 occupational groups, preserving complete occupational coverage while substantially reducing the computational cost of evaluation. We use English exclusively as the source language and consider seven target languages: Arabic, Czech, German, Greek, Icelandic, Russian, and Ukrainian. Six of these target languages are inherited from the original resource, while German is newly added in this work, yielding 1,308 masculine/feminine translation pairs per target language. We use this benchmark to evaluate both shared-task submissions and baselines for score prediction and error annotation.

However, identifying whether an evaluator exhibits gender bias is not fully captured by a single average score difference. A metric may react strongly to the gendered realization of many examples while showing little net preference because differences in opposite directions cancel out. Conversely, relatively small score differences may consistently favor the same gender across many examples. We therefore examine several complementary aspects of evaluator behavior: signed score differences to capture the direction of preference, absolute paired differences to capture sensitivity regardless of direction, and preference frequencies to measure how consistently one variant is favored. For error-annotation systems, we additionally compare how often masculine and feminine variants are predicted to be error-free.

Our contributions are threefold. First, we construct an occupation-balanced English-source subset of GAMBIT+ for the seven language pairs considered in this work, including the German extension. Second, we evaluate gender-related differences across score-prediction submissions and baselines, jointly analyzing their direction, magnitude, and frequency. Third, we extend the analysis to error-annotation systems, examining whether masculine and feminine translations differ in their likelihood of being labeled error-free. Overall, our results show that gender bias remains present in MT evaluation, but that its patterns are not fully captured by a single measure and require a more nuanced analysis of evaluator behavior.

\section{Related Work}

Gender bias in machine translation has been extensively studied across languages, model architectures, and evaluation settings \citep{savoldi-etal-2021-gender,vanmassenhove2024gender,savoldi2024gender,rescigno2020gender, paolucci2023gender,ghosh2023gender,kostikova2023gender,piazzolla2023gender}. A particularly relevant strand of this work concerns occupational gender bias, where translation systems may associate professions with stereotypical gender distributions \citep{menis2025gender, menis2025assumedpublished, menis2025gostpublished, tal2022gender,gorti2024gender}. Recent work continues to find such effects in modern MT architectures, suggesting that default gender preferences remain an open problem even as model capabilities improve \citep{manna-etal-2026-gender}.

Evaluating these biases is particularly challenging when the source does not provide enough information to determine a person's gender. Benchmarks such as WinoMT and MT-GenEval focus primarily on cases where contextual evidence identifies the appropriate gender \citep{stanovsky-etal-2019-evaluating,currey-etal-2022-mt}, whereas GAMBIT targets occupational references for which masculine and feminine realizations may both be valid \citep{menis2025assumedpublished}. This distinction has received increasing attention: recent work has explored uncertainty as a way of evaluating model behavior under gender ambiguity \citep{staliunaite-etal-2026-uncertainty}, while other studies emphasize preserving ambiguity through gender-neutral or gender-inclusive translation strategies \citep{dawkins-etal-2025-gender-neutral, savoldi-etal-2025-mind}. Our setting is complementary to these approaches, focusing specifically on systematic preferences between masculine and feminine realizations rather than evaluating neutral alternatives.

While most research has focused on biases in MT outputs, less attention has been paid to biases introduced during automatic evaluation. \citet{zaranis-etal-2025-watching} showed that quality-estimation metrics can exhibit systematic gender disparities, and GAMBIT+ extended this analysis to a multilingual, occupation-indexed challenge set \citep{filandrianos2025gambitplus}. More recently, FairQE has explored explicitly mitigating gender bias in quality estimation \citep{jang-etal-2026-fairqe}. Building on this line of work, we use GAMBIT+ to evaluate WMT 2026 metrics and extend the analysis beyond average score differences, considering complementary dimensions of evaluator preference and sensitivity.

\section{The Challenge Set}

\subsection{GAMBIT+ extension}

GAMBIT+ \citep{filandrianos2025gambitplus} provides English texts containing gender-ambiguous occupational references, together with paired target-language translations in which the relevant occupation is realized once in masculine and once in feminine form. The dataset covers the 436 four-digit occupational groups of ISCO-08 and was originally released for multiple source and target languages.

For this year's evaluation, we focus exclusively on English as the source language and consider seven target languages: Arabic, Czech, German, Greek, Icelandic, Russian, and Ukrainian. Six of these language pairs are inherited from the original GAMBIT+ release, while German is newly added in this work. We follow the same general generation procedure described in GAMBIT+ for constructing paired masculine and feminine translations, using LongCat-2.0\footnote{https://longcat.ai/} for the German data. As in the original benchmark, the two target variants are intended to preserve the same source information while differing in the gender realization of the ambiguous occupational reference and any grammatical elements that depend on it. Table~\ref{tab:german-example} shows an example.

\begin{table}[t]
\centering\small
\begin{tabularx}{\columnwidth}{@{}lX@{}}
\toprule
ISCO & 1111: Legislators \\
\midrule
Source & The Member of Parliament delivered a compelling speech about healthcare reform during yesterday's session. \\
Masculine & \underline{Der} Abgeordnete hielt w\"ahrend der gestrigen Sitzung eine \"uberzeugende Rede \"uber die Gesundheitsreform. \\
Feminine & \underline{Die} Abgeordnete hielt w\"ahrend der gestrigen Sitzung eine \"uberzeugende Rede \"uber die Gesundheitsreform. \\
\bottomrule
\end{tabularx}
\caption{An English source text with the paired German translations introduced in the GAMBIT+ extension. The underlined text marks the gender-dependent difference between the masculine and feminine variants; the remaining text is identical.}
\label{tab:german-example}
\end{table}

\subsection{Occupation-balanced subset}

To make the benchmark more practical for evaluating a larger and increasingly computationally demanding set of metrics, we construct a smaller occupation-balanced subset of GAMBIT+. We sample three English source examples for each of the 436 ISCO-08 occupational groups, yielding 1,308 source instances per language pair. This preserves complete occupational coverage while substantially reducing the size of the benchmark. Each source instance is still associated with two target translations: one masculine and one feminine.
The resulting benchmark therefore contains seven English--target language pairs, with 1,308 paired translations per target language. 

\subsection{Evaluated systems}

We evaluate the WMT 2026 Automated Translation Quality Evaluation Shared Task systems available for these seven language pairs, including both participant submissions and shared-task baselines. The analysis covers two types of evaluator output: numerical quality scores and error-span annotations.

For score prediction, each evaluator assigns a numerical quality score to the masculine and feminine translation of every source example. For error annotation, systems instead identify translation errors through spans and omission flags. These two settings provide complementary views of evaluator behavior: the former allows us to measure differences in assigned quality scores, while the latter allows us to examine whether masculine and feminine variants are treated differently in terms of predicted errors.

All analyses are performed on the complete set of 1,308 masculine/feminine pairs available for each evaluated language--system combination. We additionally verify the consistency of segment identifiers and pair alignment before computing the results. Appendix~\ref{app:coverage} reports the detailed system coverage and score ranges. Since the benchmark is stratified by occupation, the analysis below treats occupation as the primary grouping unit when assessing the stability and consistency of gender-related score differences.

\section{Analysis Protocol}
\subsection{Direction, magnitude and frequency}
Let $q_{k\ell i}^{M}$ and $q_{k\ell i}^{F}$ be the scores for metric $k$, target $\ell$ and sample $i$, for the masculine and feminine translations respectively. We define:
\begin{align}
d_{k\ell i}&=q_{k\ell i}^{M}-q_{k\ell i}^{F},\\
S_{k\ell}&=\frac{1}{n_\ell}\sum_i d_{k\ell i},\label{eq:signed}\\
A_{k\ell}&=\frac{1}{n_\ell}\sum_i |d_{k\ell i}|.\label{eq:absolute}
\end{align}
Positive $S$ values indicate higher masculine scores, while $A$ measures sensitivity regardless of direction. In particular, $A\ne|S|$ in general: opposing differences cancel in $S$ but not in $A$.

For comparability with the range-based analysis of GAMBIT+, we also report
\begin{equation}
S^{\%}_{k\ell}=100S_{k\ell}/R_k,\quad
A^{\%}_{k\ell}=100A_{k\ell}/R_k,
\end{equation}
where $R_k$ is the maximum minus minimum \emph{individual} score over both variants and all available target languages for that submission. These percentages express the differences as a fraction of the observed range. They remain sensitive to outliers and coverage.

We additionally calculate masculine-win, feminine-win and tie rates, with all pairs considered in the denominator. The preference balance then corresponds to $100[\Pr(d>0)-\Pr(d<0)]$ percentage points. For $A>0$, the cancellation ratio $1-|S|/A$ quantifies the discrepancy between net and absolute effects. 

\subsection{Aggregation, confidence intervals, and significance testing}

Each target language contains three examples for each of the 436 ISCO-08 occupational groups. Since every occupation is represented equally, averaging over all 1,308 translation pairs is equivalent to first averaging within each occupation and then averaging over occupations. For multilingual summaries, we compute macro-averages across languages, giving each language equal weight. We report averages over all available languages for each system, and use a fixed six-language panel (AR, CS, DE, IS, RU, and UK) when comparing the 22 systems with common coverage.

To assess the stability of the aggregate effects across occupations, we use an occupation-block bootstrap. We resample the 436 occupations with replacement, retaining the three paired examples belonging to each selected occupation, and recompute the signed and absolute differences over 5,000 bootstrap replicates. For multilingual summaries, all observations associated with the same occupation across the included languages remain in the same block. We report percentile 95\% intervals while keeping the observed normalization ranges fixed. Because the benchmark already covers all 436 ISCO-08 groups, these intervals should be interpreted as sensitivity to changes in the relative weighting of occupations, rather than as sampling uncertainty over unseen occupations or languages.

For each system--language run, we test whether the mean signed difference across the 436 occupations differs from zero using a two-sided one-sample $t$-test. We correct the resulting 148 $p$-values for multiple comparisons using the Benjamini--Hochberg procedure \citep{benjamini1995fdr}. A significant result indicates a consistent directional preference across occupations; the sign of the mean determines whether masculine or feminine variants receive higher scores on average.

\section{Score-Prediction Results}
\begin{table*}[t]
\centering
\scriptsize
\setlength{\tabcolsep}{3pt}
\begin{tabular}{lccccccc|c}
\toprule
Submission & AR & CS & DE & EL & IS & RU & UK & Macro-6 \\
\midrule
AEGIS & +0.66/1.20 & -0.03/1.79 & +0.26/0.57 & +0.45/0.74 & +0.61/1.96 & +0.67/1.47 & +0.19/1.13 & +0.39/1.35 \\
$^{\dagger}$COMETKiwi22 & +0.91/1.38 & +0.52/0.94 & +0.20/0.57 & +0.68/0.82 & +1.62/1.95 & +1.11/1.37 & +0.46/0.95 & +0.80/1.19 \\
Cohere CAT+ ensemble & +3.56/5.07 & +2.68/4.63 & +1.08/4.10 & +2.19/3.48 & +2.59/5.07 & +5.13/6.65 & +2.12/4.51 & +2.86/5.01 \\
FACET\_869543 & +2.21/3.99 & +2.18/4.16 & +0.38/4.06 & +2.32/4.72 & +1.77/4.11 & +3.09/5.04 & +1.37/3.81 & +1.83/4.19 \\
FACET\_869546 & +2.26/4.29 & +2.17/4.33 & +0.43/4.24 & +2.39/4.81 & +1.81/4.17 & +3.16/5.22 & +1.40/3.86 & +1.87/4.35 \\
$^{\dagger}$Gemini-3.6-Flash & +3.02/5.25 & +1.57/3.59 & +0.08/3.01 & -- & +2.19/4.06 & +3.24/4.39 & +1.17/3.14 & +1.88/3.91 \\
$^{\dagger}$Gemma 4 & +8.91/14.17 & +3.68/5.24 & +0.84/3.46 & -- & +4.62/6.33 & +5.70/6.71 & +3.71/5.35 & +4.58/6.88 \\
$^{\dagger}$Gemma 4 RB & +8.93/14.25 & +3.61/5.22 & +0.74/3.27 & -- & +4.62/6.33 & +5.48/6.65 & +3.71/5.35 & +4.52/6.85 \\
Lexicala (Majority-Vote) & +11.47/14.58 & -- & -- & -- & -- & -- & -- & -- \\
Lexicala (STAPLE) & +12.11/15.72 & -- & -- & -- & -- & -- & -- & -- \\
MQM-LLM & +3.51/5.04 & +2.82/4.68 & +2.08/4.46 & +2.70/4.18 & +4.37/6.56 & +4.93/6.08 & +4.11/5.67 & +3.64/5.41 \\
MQM-LLM CA & +3.64/5.22 & +3.07/5.30 & +2.08/5.17 & +3.29/4.85 & +4.85/7.20 & +5.24/6.74 & +4.48/6.36 & +3.89/6.00 \\
$^{\dagger}$Qwen 3.6 & +3.89/7.59 & +3.98/6.92 & +2.09/6.15 & -- & +4.05/8.50 & +5.77/7.84 & +4.58/7.03 & +4.06/7.34 \\
$^{\dagger}$Qwen 3.6 RB & +3.80/7.52 & +3.95/6.92 & +1.72/5.47 & -- & +4.05/8.50 & +5.82/7.95 & +4.58/7.03 & +3.99/7.23 \\
Vertical\_870257 & -0.24/2.67 & +0.04/1.79 & +0.07/1.31 & +1.23/1.68 & +0.46/1.54 & +0.63/1.39 & +0.95/1.82 & +0.32/1.75 \\
Vertical\_870554 & -0.46/2.41 & -0.11/2.64 & -0.34/2.51 & +2.34/3.19 & +0.33/2.49 & +1.06/2.55 & +0.82/1.76 & +0.22/2.39 \\
bytepop\_pro10 & +3.32/4.56 & +0.96/2.33 & -0.08/1.79 & +3.76/5.59 & +4.44/5.85 & +2.94/3.42 & +2.54/4.26 & +2.35/3.70 \\
bytepop\_pro9 & +3.09/4.16 & +0.96/2.33 & -0.08/1.79 & +3.76/5.59 & +3.71/5.25 & +2.94/3.42 & +2.54/4.26 & +2.19/3.54 \\
cuni-v14-regression & -- & +8.36/11.58 & -- & -- & -- & -- & -- & -- \\
fluency2-gemini35-esa & -0.61/14.04 & -0.92/14.25 & -2.59/8.93 & +1.20/16.30 & +1.35/13.17 & +3.40/12.11 & -1.97/13.63 & -0.22/12.69 \\
gemba-poly & +0.84/1.49 & +0.17/0.93 & -0.07/1.35 & +0.12/1.24 & -0.00/1.05 & +1.01/2.04 & +0.10/1.05 & +0.34/1.32 \\
$^{\dagger}$xCOMET XL RB & +2.60/4.77 & +2.11/3.86 & +0.63/1.51 & -- & +3.98/4.95 & +3.59/4.37 & +2.03/3.30 & +2.49/3.79 \\
$^{\dagger}$xCOMET XL RF & +2.60/4.77 & +2.11/3.86 & +0.63/1.51 & -- & +3.98/4.95 & +3.59/4.37 & +2.03/3.30 & +2.49/3.79 \\
$^{\dagger}$xCOMET XXL RB & +4.93/6.60 & +5.00/6.56 & +0.95/1.93 & -- & +6.71/7.70 & +8.97/9.94 & +7.94/9.32 & +5.75/7.01 \\
$^{\dagger}$xCOMET XXL RF & +4.93/6.60 & +5.00/6.56 & +0.95/1.93 & -- & +6.71/7.70 & +8.97/9.94 & +7.94/9.32 & +5.75/7.01 \\
\bottomrule
\end{tabular}
\caption{Signed / mean absolute paired differences ($S^{\%}/A^{\%}$), in percent of each evaluator\textquotesingle s observed score range. Positive signed values indicate higher masculine scores. Each language has 1,308 pairs. Macro-6 equally averages AR, CS, DE, IS, RU and UK; -- denotes missing coverage, never zero. RB/RF retain returned reference-based/reference-free labels, and CA abbreviates Confidence Aware; inference configurations were not independently verified. Native scales appear in Appendix~\ref{app:coverage}. $^{\dagger}$ denotes a WMT 2026 organizer baseline; unmarked systems
are participant submissions.} 
\label{tab:metric-language-2026}
\end{table*}

\subsection{Systematic preference and cancellation}
Table~\ref{tab:metric-language-2026} reports both normalized summaries for every Task~2 (Quality Score Prediction) system. After Benjamini--Hochberg correction at $q=0.05$, 131 of the 148 system--language runs show a statistically detectable directional difference. Of these, 126 favor masculine variants on average and five favor feminine variants.
The five significant negative runs are \textsc{Vertical\_870257} on Arabic, \textsc{Vertical\_870554} on Arabic and German, and \textsc{fluency2-gemini35-esa} on German and Ukrainian. Significance should also be read together with magnitude: a consistent but small preference and a large, heterogeneous effect have different implications.

\begin{figure*}[t]
\centering
\begin{tikzpicture}[x=0.62cm,y=0.245cm]
\definecolor{signedblue}{RGB}{43,86,130}
\definecolor{absorange}{RGB}{190,112,28}
\node[anchor=east,font=\scriptsize] at (-1.1,0) {COMETKiwi22};
\draw[black!20,line width=1.2pt] (0.804229,0) -- (1.193266,0);
\draw[signedblue,line width=.5pt] (0.731317,0) -- (0.882811,0);
\draw[signedblue] (0.731317,-0.17) -- (0.731317,0.17);
\draw[signedblue] (0.882811,-0.17) -- (0.882811,0.17);
\fill[signedblue] (0.804229,0) circle[radius=1.5pt];
\node[fill=absorange,minimum size=3pt,inner sep=0pt] at (1.193266,0) {};
\node[anchor=east,font=\scriptsize] at (-1.1,1) {gemba-poly};
\draw[black!20,line width=1.2pt] (0.342125,1) -- (1.319190,1);
\draw[signedblue,line width=.5pt] (0.240953,1) -- (0.448273,1);
\draw[signedblue] (0.240953,0.83) -- (0.240953,1.17);
\draw[signedblue] (0.448273,0.83) -- (0.448273,1.17);
\fill[signedblue] (0.342125,1) circle[radius=1.5pt];
\node[fill=absorange,minimum size=3pt,inner sep=0pt] at (1.319190,1) {};
\node[anchor=east,font=\scriptsize] at (-1.1,2) {AEGIS};
\draw[black!20,line width=1.2pt] (0.394769,2) -- (1.352179,2);
\draw[signedblue,line width=.5pt] (0.324136,2) -- (0.468667,2);
\draw[signedblue] (0.324136,1.83) -- (0.324136,2.17);
\draw[signedblue] (0.468667,1.83) -- (0.468667,2.17);
\fill[signedblue] (0.394769,2) circle[radius=1.5pt];
\node[fill=absorange,minimum size=3pt,inner sep=0pt] at (1.352179,2) {};
\node[anchor=east,font=\scriptsize] at (-1.1,3) {Vertical\_870257};
\draw[black!20,line width=1.2pt] (0.317637,3) -- (1.752991,3);
\draw[signedblue,line width=.5pt] (0.243659,3) -- (0.392823,3);
\draw[signedblue] (0.243659,2.83) -- (0.243659,3.17);
\draw[signedblue] (0.392823,2.83) -- (0.392823,3.17);
\fill[signedblue] (0.317637,3) circle[radius=1.5pt];
\node[fill=absorange,minimum size=3pt,inner sep=0pt] at (1.752991,3) {};
\node[anchor=east,font=\scriptsize] at (-1.1,4) {Vertical\_870554};
\draw[black!20,line width=1.2pt] (0.217848,4) -- (2.393872,4);
\draw[signedblue,line width=.5pt] (0.120599,4) -- (0.312977,4);
\draw[signedblue] (0.120599,3.83) -- (0.120599,4.17);
\draw[signedblue] (0.312977,3.83) -- (0.312977,4.17);
\fill[signedblue] (0.217848,4) circle[radius=1.5pt];
\node[fill=absorange,minimum size=3pt,inner sep=0pt] at (2.393872,4) {};
\node[anchor=east,font=\scriptsize] at (-1.1,5) {bytepop\_pro9};
\draw[black!20,line width=1.2pt] (2.193841,5) -- (3.535880,5);
\draw[signedblue,line width=.5pt] (1.985313,5) -- (2.401936,5);
\draw[signedblue] (1.985313,4.83) -- (1.985313,5.17);
\draw[signedblue] (2.401936,4.83) -- (2.401936,5.17);
\fill[signedblue] (2.193841,5) circle[radius=1.5pt];
\node[fill=absorange,minimum size=3pt,inner sep=0pt] at (3.535880,5) {};
\node[anchor=east,font=\scriptsize] at (-1.1,6) {bytepop\_pro10};
\draw[black!20,line width=1.2pt] (2.352510,6) -- (3.702340,6);
\draw[signedblue,line width=.5pt] (2.130146,6) -- (2.581117,6);
\draw[signedblue] (2.130146,5.83) -- (2.130146,6.17);
\draw[signedblue] (2.581117,5.83) -- (2.581117,6.17);
\fill[signedblue] (2.352510,6) circle[radius=1.5pt];
\node[fill=absorange,minimum size=3pt,inner sep=0pt] at (3.702340,6) {};
\node[anchor=east,font=\scriptsize] at (-1.1,7) {xCOMET XL RB};
\draw[black!20,line width=1.2pt] (2.489543,7) -- (3.792614,7);
\draw[signedblue,line width=.5pt] (2.291555,7) -- (2.685592,7);
\draw[signedblue] (2.291555,6.83) -- (2.291555,7.17);
\draw[signedblue] (2.685592,6.83) -- (2.685592,7.17);
\fill[signedblue] (2.489543,7) circle[radius=1.5pt];
\node[fill=absorange,minimum size=3pt,inner sep=0pt] at (3.792614,7) {};
\node[anchor=east,font=\scriptsize] at (-1.1,8) {xCOMET XL RF};
\draw[black!20,line width=1.2pt] (2.489543,8) -- (3.792614,8);
\draw[signedblue,line width=.5pt] (2.291784,8) -- (2.691854,8);
\draw[signedblue] (2.291784,7.83) -- (2.291784,8.17);
\draw[signedblue] (2.691854,7.83) -- (2.691854,8.17);
\fill[signedblue] (2.489543,8) circle[radius=1.5pt];
\node[fill=absorange,minimum size=3pt,inner sep=0pt] at (3.792614,8) {};
\node[anchor=east,font=\scriptsize] at (-1.1,9) {Gemini-3.6-Flash};
\draw[black!20,line width=1.2pt] (1.879332,9) -- (3.906091,9);
\draw[signedblue,line width=.5pt] (1.708827,9) -- (2.054285,9);
\draw[signedblue] (1.708827,8.83) -- (1.708827,9.17);
\draw[signedblue] (2.054285,8.83) -- (2.054285,9.17);
\fill[signedblue] (1.879332,9) circle[radius=1.5pt];
\node[fill=absorange,minimum size=3pt,inner sep=0pt] at (3.906091,9) {};
\node[anchor=east,font=\scriptsize] at (-1.1,10) {FACET\_869543};
\draw[black!20,line width=1.2pt] (1.832739,10) -- (4.194881,10);
\draw[signedblue,line width=.5pt] (1.657664,10) -- (2.013555,10);
\draw[signedblue] (1.657664,9.83) -- (1.657664,10.17);
\draw[signedblue] (2.013555,9.83) -- (2.013555,10.17);
\fill[signedblue] (1.832739,10) circle[radius=1.5pt];
\node[fill=absorange,minimum size=3pt,inner sep=0pt] at (4.194881,10) {};
\node[anchor=east,font=\scriptsize] at (-1.1,11) {FACET\_869546};
\draw[black!20,line width=1.2pt] (1.873089,11) -- (4.352337,11);
\draw[signedblue,line width=.5pt] (1.698947,11) -- (2.056332,11);
\draw[signedblue] (1.698947,10.83) -- (1.698947,11.17);
\draw[signedblue] (2.056332,10.83) -- (2.056332,11.17);
\fill[signedblue] (1.873089,11) circle[radius=1.5pt];
\node[fill=absorange,minimum size=3pt,inner sep=0pt] at (4.352337,11) {};
\node[anchor=east,font=\scriptsize] at (-1.1,12) {Cohere CAT+ ensemble};
\draw[black!20,line width=1.2pt] (2.859731,12) -- (5.005288,12);
\draw[signedblue,line width=.5pt] (2.538162,12) -- (3.192805,12);
\draw[signedblue] (2.538162,11.83) -- (2.538162,12.17);
\draw[signedblue] (3.192805,11.83) -- (3.192805,12.17);
\fill[signedblue] (2.859731,12) circle[radius=1.5pt];
\node[fill=absorange,minimum size=3pt,inner sep=0pt] at (5.005288,12) {};
\node[anchor=east,font=\scriptsize] at (-1.1,13) {MQM-LLM};
\draw[black!20,line width=1.2pt] (3.635640,13) -- (5.414034,13);
\draw[signedblue,line width=.5pt] (3.389894,13) -- (3.887559,13);
\draw[signedblue] (3.389894,12.83) -- (3.389894,13.17);
\draw[signedblue] (3.887559,12.83) -- (3.887559,13.17);
\fill[signedblue] (3.635640,13) circle[radius=1.5pt];
\node[fill=absorange,minimum size=3pt,inner sep=0pt] at (5.414034,13) {};
\node[anchor=east,font=\scriptsize] at (-1.1,14) {MQM-LLM CA};
\draw[black!20,line width=1.2pt] (3.894169,14) -- (5.997584,14);
\draw[signedblue,line width=.5pt] (3.633641,14) -- (4.139868,14);
\draw[signedblue] (3.633641,13.83) -- (3.633641,14.17);
\draw[signedblue] (4.139868,13.83) -- (4.139868,14.17);
\fill[signedblue] (3.894169,14) circle[radius=1.5pt];
\node[fill=absorange,minimum size=3pt,inner sep=0pt] at (5.997584,14) {};
\node[anchor=east,font=\scriptsize] at (-1.1,15) {Gemma 4 RB};
\draw[black!20,line width=1.2pt] (4.517329,15) -- (6.845821,15);
\draw[signedblue,line width=.5pt] (4.172340,15) -- (4.870846,15);
\draw[signedblue] (4.172340,14.83) -- (4.172340,15.17);
\draw[signedblue] (4.870846,14.83) -- (4.870846,15.17);
\fill[signedblue] (4.517329,15) circle[radius=1.5pt];
\node[fill=absorange,minimum size=3pt,inner sep=0pt] at (6.845821,15) {};
\node[anchor=east,font=\scriptsize] at (-1.1,16) {Gemma 4};
\draw[black!20,line width=1.2pt] (4.576835,16) -- (6.878058,16);
\draw[signedblue,line width=.5pt] (4.241319,16) -- (4.916030,16);
\draw[signedblue] (4.241319,15.83) -- (4.241319,16.17);
\draw[signedblue] (4.916030,15.83) -- (4.916030,16.17);
\fill[signedblue] (4.576835,16) circle[radius=1.5pt];
\node[fill=absorange,minimum size=3pt,inner sep=0pt] at (6.878058,16) {};
\node[anchor=east,font=\scriptsize] at (-1.1,17) {xCOMET XXL RB};
\draw[black!20,line width=1.2pt] (5.750325,17) -- (7.008352,17);
\draw[signedblue,line width=.5pt] (5.363774,17) -- (6.121662,17);
\draw[signedblue] (5.363774,16.83) -- (5.363774,17.17);
\draw[signedblue] (6.121662,16.83) -- (6.121662,17.17);
\fill[signedblue] (5.750325,17) circle[radius=1.5pt];
\node[fill=absorange,minimum size=3pt,inner sep=0pt] at (7.008352,17) {};
\node[anchor=east,font=\scriptsize] at (-1.1,18) {xCOMET XXL RF};
\draw[black!20,line width=1.2pt] (5.750325,18) -- (7.008352,18);
\draw[signedblue,line width=.5pt] (5.364803,18) -- (6.130489,18);
\draw[signedblue] (5.364803,17.83) -- (5.364803,18.17);
\draw[signedblue] (6.130489,17.83) -- (6.130489,18.17);
\fill[signedblue] (5.750325,18) circle[radius=1.5pt];
\node[fill=absorange,minimum size=3pt,inner sep=0pt] at (7.008352,18) {};
\node[anchor=east,font=\scriptsize] at (-1.1,19) {Qwen 3.6 RB};
\draw[black!20,line width=1.2pt] (3.987385,19) -- (7.233308,19);
\draw[signedblue,line width=.5pt] (3.667896,19) -- (4.297799,19);
\draw[signedblue] (3.667896,18.83) -- (3.667896,19.17);
\draw[signedblue] (4.297799,18.83) -- (4.297799,19.17);
\fill[signedblue] (3.987385,19) circle[radius=1.5pt];
\node[fill=absorange,minimum size=3pt,inner sep=0pt] at (7.233308,19) {};
\node[anchor=east,font=\scriptsize] at (-1.1,20) {Qwen 3.6};
\draw[black!20,line width=1.2pt] (4.061417,20) -- (7.339195,20);
\draw[signedblue,line width=.5pt] (3.740568,20) -- (4.378832,20);
\draw[signedblue] (3.740568,19.83) -- (3.740568,20.17);
\draw[signedblue] (4.378832,19.83) -- (4.378832,20.17);
\fill[signedblue] (4.061417,20) circle[radius=1.5pt];
\node[fill=absorange,minimum size=3pt,inner sep=0pt] at (7.339195,20) {};
\node[anchor=east,font=\scriptsize] at (-1.1,21) {fluency2-gemini35-esa};
\draw[black!20,line width=1.2pt] (-0.220878,21) -- (12.688922,21);
\draw[signedblue,line width=.5pt] (-0.903571,21) -- (0.476950,21);
\draw[signedblue] (-0.903571,20.83) -- (-0.903571,21.17);
\draw[signedblue] (0.476950,20.83) -- (0.476950,21.17);
\fill[signedblue] (-0.220878,21) circle[radius=1.5pt];
\node[fill=absorange,minimum size=3pt,inner sep=0pt] at (12.688922,21) {};
\draw[black!40] (0,-.7) -- (0,21.6);
\draw (0,-.65) -- (0,-.85);
\node[below,font=\scriptsize] at (0,-.85) {0};
\draw (2,-.65) -- (2,-.85);
\node[below,font=\scriptsize] at (2,-.85) {2};
\draw (4,-.65) -- (4,-.85);
\node[below,font=\scriptsize] at (4,-.85) {4};
\draw (6,-.65) -- (6,-.85);
\node[below,font=\scriptsize] at (6,-.85) {6};
\draw (8,-.65) -- (8,-.85);
\node[below,font=\scriptsize] at (8,-.85) {8};
\draw (10,-.65) -- (10,-.85);
\node[below,font=\scriptsize] at (10,-.85) {10};
\draw (12,-.65) -- (12,-.85);
\node[below,font=\scriptsize] at (12,-.85) {12};
\node[font=\small] at (6,-3.1) {Difference (\% of observed metric range)};
\node[font=\scriptsize,anchor=west,text=signedblue] at (5.8,1.2) {$\bullet$ Signed mean (95\% interval)};
\node[font=\scriptsize,anchor=west,text=absorange] at (5.8,0) {$\blacksquare$ Mean absolute difference};
\end{tikzpicture}
\caption{Signed and mean absolute gender-related score differences on the common six-language set across 22 qualifying evaluators. The signed difference captures the direction of preference, while the mean absolute difference captures its magnitude regardless of direction. Whiskers show percentile 95\% intervals from 5,000 resamples of the 436 ISCO blocks, keeping paired observations and included languages together. Values closer to zero indicate smaller gender-related differences.}
\label{fig:direction-sensitivity}
\end{figure*}
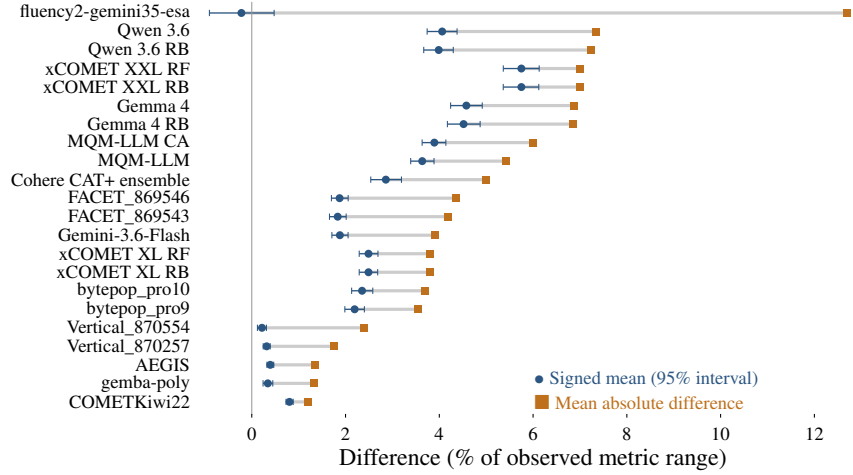

Figure~\ref{fig:direction-sensitivity} compares the signed and mean absolute differences on the common six-language set. The contrast between the two quantities is particularly clear for \textsc{fluency2-gemini35-esa}: although its mean absolute difference is large ($A=5.7234$, $A^{\%}=12.69$), its signed mean is close to zero ($S=-0.0996$, $S^{\%}=-0.22$). This indicates that substantial gender-related score changes occur in both directions and largely cancel in the signed average. Qwen 3.6, by comparison, has $S^{\%}=4.06$ and $A^{\%}=7.34$, while \textsc{xCOMET XXL} reference-free has $S^{\%}=5.75$ and $A^{\%}=7.01$, indicating a more consistently positive preference relative to the overall magnitude of the differences.

\textsc{COMETKiwi22} \citep{rei-etal-2022-cometkiwi} illustrates a different pattern. Its six-language mean absolute difference is small ($A=0.00674$, corresponding to 1.19\% of its observed range), yet it assigns higher scores to masculine variants in 78.1\% of pairs, compared with 20.8\% for feminine variants and 1.1\% ties. Thus, a small average magnitude does not necessarily imply balanced preference frequencies. Conversely, \textsc{gemba-poly} produces ties for 61.8\% of pairs. 

\subsection{Language and occupation differences}

To compare target languages using the same set of evaluators, we average the normalized differences over the 13 systems available for all seven languages. The mean absolute difference, $A^{\%}$, ranges from 3.14 in German to 4.64 in Icelandic. Russian shows the largest average signed difference ($S^{\%}=2.72$), while German is much closer to zero ($S^{\%}=0.26$). 

The direction of the preference can also vary across evaluators. For German, for example, Qwen 3.6 has a positive signed difference ($S^{\%}=2.09$), whereas \textsc{fluency2-gemini35-esa} has a negative one ($S^{\%}=-2.59$). Thus, even within the same target language, evaluators do not always favor the same gender. This suggests that the observed differences are more strongly associated with the evaluator than with the target language.

We also examine differences across occupations. Table~\ref{tab:multilingual-occupations} reports the five occupations with the largest feminine- and masculine-preferring signed differences, based on the mean normalized signed difference $S^{\%}$ across the seven target languages. To make these comparisons consistent, all language-level values are averaged over the same 13 systems with coverage for all seven languages.

Several of the strongest differences align with familiar occupational gender stereotypes. Midwifery professionals and child care workers receive higher feminine scores on average, whereas plumbers, hunters, and trappers receive higher masculine scores. This pattern is broadly consistent with observations in our previous GAMBIT+ evaluation. However, the association is not uniform across occupations or languages: for example, Fashion and Other Models shows a feminine preference overall despite positive differences in Arabic and Czech. We therefore treat these results as descriptive evidence of occupation-level variation rather than as a direct measure of gender stereotypes. 

\begin{table*}[t]
\centering
\small
\setlength{\tabcolsep}{2.2pt}
\begin{tabularx}{\textwidth}{@{}lXrrrrrrr|r@{}}
\toprule
ISCO & Occupation & AR & CS & DE & EL & IS & RU & UK & Mean \\
\midrule
2222 & Midwifery Professionals & -7.27 & -7.43 & -4.45 & -9.63 & -11.79 & -5.69 & -7.87 & -7.73 \\
3222 & Midwifery Associate Professionals & -3.09 & -4.82 & -3.42 & -5.70 & -10.23 & -5.66 & -1.85 & -4.97 \\
5311 & Child Care Workers & -4.24 & -0.27 & -0.36 & -3.03 & -5.11 & -8.31 & -2.57 & -3.41 \\
5241 & Fashion and Other Models & +4.41 & +0.81 & -3.25 & -4.86 & -3.56 & -5.34 & -10.69 & -3.21 \\
5151 & Cleaning and Housekeeping Supervisors in Offices, Hotels and Other Establishments & -2.78 & -4.49 & -2.51 & -2.72 & -3.79 & -3.44 & -0.86 & -2.94 \\
\midrule
2636 & Religious Professionals & +8.14 & +3.28 & +4.85 & +8.22 & +3.34 & +4.44 & +8.57 & +5.84 \\
4213 & Pawnbrokers and Money-lenders & +6.77 & +3.35 & +1.14 & +12.32 & +5.39 & +8.18 & +5.62 & +6.11 \\
4414 & Scribes and Related Workers & +4.09 & +2.60 & +3.42 & +9.90 & +5.63 & +11.94 & +5.53 & +6.16 \\
6224 & Hunters and Trappers & +6.81 & +6.11 & +6.52 & +9.26 & +3.94 & +7.05 & +7.40 & +6.73 \\
7126 & Plumbers and Pipe Fitters & +15.34 & +3.73 & +1.20 & +11.83 & +4.09 & +7.00 & +5.55 & +6.96 \\
\bottomrule
\end{tabularx}
\caption{Occupation-level extremes across the seven target languages, using the fixed panel of 13 Task~2 systems available for all languages. Each language column reports the mean normalized signed difference $S^{\%}$ across these systems, and the final column gives the macro-average across languages. Negative values indicate higher scores for feminine variants and positive values higher scores for masculine variants.}
\label{tab:multilingual-occupations}
\end{table*}

\begin{table*}[!t]
\centering
\scriptsize
\setlength{\tabcolsep}{3pt}
\begin{tabular}{lrrrrrrr}
\toprule
Submission & AR & CS & DE & EL & IS & RU & UK \\
\midrule
Cohere CAT+ & +8.7 & +4.2 & +2.2 & +4.0 & +3.7 & +3.1 & -0.5 \\
FACET\_869532 & +4.3 & +9.3 & +1.3 & +4.8 & +8.6 & +10.1 & +20.3 \\
FACET\_869533 & +4.3 & +9.6 & +1.2 & +4.9 & +8.6 & +10.2 & +20.6 \\
FACET\_876357 & +4.3 & +9.3 & +1.3 & +4.8 & +8.6 & +10.1 & +20.3 \\
FACET\_876365 & +4.3 & +9.6 & +1.2 & +4.9 & +8.6 & +10.2 & +20.6 \\
$^{\dagger}$Gemini-3.6-Flash  & +16.4 & +7.0 & +0.3 & -- & +7.3 & +13.4 & +4.5 \\
$^{\dagger}$Gemma 4 & +15.1 & +19.1 & +3.1 & -- & +16.1 & +27.8 & +18.2 \\
$^{\dagger}$Gemma 4 RB  & +15.1 & +19.3 & +4.1 & -- & +16.1 & +28.4 & +18.2 \\
Lexicala (Majority-Vote) & +30.0 & -- & -- & -- & -- & -- & -- \\
Lexicala (Majority-Vote)\_876237 & +30.0 & -- & -- & -- & -- & -- & -- \\
Lexicala (Majority-Vote)\_876371 & +30.0 & -- & -- & -- & -- & -- & -- \\
Lexicala (STAPLE) & +30.3 & -- & -- & -- & -- & -- & -- \\
Lexicala (STAPLE)\_876238 & +30.3 & -- & -- & -- & -- & -- & -- \\
Lexicala (STAPLE)\_876370 & +30.3 & -- & -- & -- & -- & -- & -- \\
$^{\dagger}$Qwen 3.6 & +7.9 & +10.3 & +6.3 & -- & +4.7 & +12.1 & +9.7 \\
$^{\dagger}$Qwen 3.6 RB & +8.0 & +10.2 & +6.0 & -- & +4.7 & +13.5 & +9.7 \\
QwenSpan & -- & -- & +2.1 & -- & -- & -- & -- \\
XCOMOffset & -- & -- & -- & -- & +1.2 & +11.9 & +3.3 \\
XCOMOffset\_875860 & -- & -- & -- & -- & +1.2 & +11.9 & +3.3 \\
XCOMOffset\_876378 & -- & -- & -- & -- & -0.2 & +0.0 & +0.0 \\
XCOMOffset\_876381 & -- & -- & -- & -- & +1.2 & +11.9 & +3.3 \\
cuni-cat-v4 & -- & +21.1 & -- & -- & -- & -- & -- \\
cuni-v14 & -- & +21.3 & -- & -- & -- & -- & -- \\
fluency2-gemini35-spans & +0.8 & -0.5 & -2.9 & +0.4 & +2.7 & +4.7 & -3.5 \\
fluencyscore-multispan-v2 & +0.8 & -0.5 & -2.9 & +0.4 & +2.7 & +4.7 & -3.5 \\
fluencyscore-multispan-v2.1 & +0.8 & -0.5 & -2.9 & +0.4 & +2.7 & +4.7 & -3.5 \\
 $^{\dagger}$xCOMET XL RF & +2.0 & +4.1 & +1.7 & -- & +3.7 & +10.7 & +5.6 \\
 $^{\dagger}$xCOMET XXL RB  & +6.7 & +7.8 & +2.1 & -- & +5.4 & +17.0 & +11.6 \\
$^{\dagger}$xCOMET XXL RF & +6.7 & +7.8 & +2.1 & -- & +5.4 & +17.0 & +11.6 \\
\bottomrule
\end{tabular}
\caption{Task~1 differences in predicted error-free rate, in percentage points (masculine minus feminine). Positive values indicate that masculine variants are more often predicted as error-free, while negative values indicate a higher error-free rate for feminine variants. $^{\dagger}$ denotes a baseline; unmarked systems are participant submissions.}
\label{tab:task1-full}
\end{table*}

\section{Error-Annotation Results}

Task~1 provides a complementary view of gender-related evaluator behavior by allowing us to examine whether masculine and feminine variants differ in how often they are predicted to be error-free. We define a translation as error-free when the system predicts neither an error span nor an omission flag. This closely parallels Task~3 of the shared task, which focuses on the detection of error-free segments. For each system and language, we compute the difference between the masculine and feminine error-free rates, in percentage points; positive values indicate that masculine variants are more often predicted as error-free.

Table~\ref{tab:task1-full} reports the results across all available systems and languages. The overall pattern is strongly skewed toward masculine variants: most system--language combinations have positive values, indicating that masculine translations are more often judged error-free than their feminine counterparts. In several cases the differences are substantial, reaching 27.8 percentage points for \textsc{Gemma 4} in Russian and around 30 points for the \textsc{Lexicala} variants in Arabic.

At the same time, the magnitude of the effect varies considerably across both systems and languages, and the dominant direction is not universal. For example, \textsc{fluency2-gemini35-spans} favors feminine variants in Czech, German, and Ukrainian, while showing positive differences in the remaining languages. More generally, systems often show similar tendencies across several languages, but different evaluators can behave very differently on the same language. This mirrors the Task~2 results and suggests that the observed gender-related differences are strongly shaped by the evaluator itself rather than a uniform language-level effect.


\section{Conclusion}

In this work, we revisited gender bias in MT evaluation using an occupation-balanced subset of GAMBIT+, covering seven English-source language pairs and extending the benchmark with German. Across WMT 2026 score-prediction systems, masculine translations are generally favored: 126 of the 131 statistically detectable system--language differences are positive. However, the strength and form of this preference vary substantially across evaluators. Signed differences, absolute differences, and preference frequencies often capture different aspects of the same behavior, while the error-annotation analysis shows a similar overall tendency for masculine variants to be predicted as error-free more often, with important exceptions.

Overall, our results show that gender-related bias remains present in modern MT evaluation systems, but cannot be adequately characterized by a single aggregate measure. The observed effects vary across evaluators, languages, and occupations, with some of the strongest occupation-level differences aligning with familiar gender stereotypes. We therefore recommend assessing evaluator bias through complementary measures of direction, magnitude, and consistency. The occupation-balanced GAMBIT+ subset introduced here provides a more computationally practical resource for such analyses while preserving complete ISCO-08 occupational coverage.

\section*{Limitations}

German was generated separately from the inherited GAMBIT+ targets, and the language-specific subsets do not necessarily contain the same English contexts. In addition, each occupation--language combination contains only three examples, so occupation-level differences should be interpreted descriptively rather than as stable estimates.

The analysis also has certain methodological limitations. Range-based normalization is sensitive to extreme scores, and small gender-related differences should not be interpreted as evidence of overall evaluator quality. We also observe only one returned run per system and therefore cannot separate systematic effects from stochastic variation. Finally, our evaluation considers only masculine and feminine variants; gender-neutral or inclusive alternatives are outside the scope of the submitted benchmark. Accordingly, we do not draw causal conclusions about language effects, occupation-specific stereotypes, or evaluator fairness beyond the pairs and systems studied here.

\section*{Acknowledgments}
O. Menis Mastromichalakis and G. Attanasio were supported by the project DECOLLAGE (ERC-2022-CoG 101088763), and FCT/MECI through national funds and co-funded EU funds under UID/50008: Instituto de Telecomunicações. G. Filandrianos and C. Zerva were partly supported within the framework of the Pharos AI Factory project, funded by the European High-Performance Computing Joint Undertaking (EuroHPC JU) under Grant Agreement No. 101234269 as part of Horizon Europe and by the Greek Public Investments Program programme. G. Attanasio was additionally supported by the AMALIA project under Measure RE-C05-i08 of the Portuguese national Programa de Recuperação e Resiliência, and by the Portuguese Recovery and Resilience Plan through project C645008882-00000055 (Center for Responsible AI). C. Zerva was additionally supported by a Google Research Scholar award. We would also like to thank Manuel Lardelli for the help in reviewing selected German masculine and feminine translations for quality and validity.



\bibliography{custom}
\clearpage

\appendix
\section{Coverage, Native Scales and Macro Results}\label{app:coverage}
Table~\ref{tab:native-scales} gives the interval of individual scores observed across each submission's available languages and the corresponding range $R_k$. Available-language means use all returned targets for that submission. Six-language means use only AR, CS, DE, IS, RU and UK.

Tables and figures abbreviate reference-based/free as RB/RF, Confidence Aware as CA, and Lexicala-QE-Ensemble as Lexicala. The unsuffixed Gemma 4 and Qwen 3.6 names refer to their returned ``Reasoning Baseline'' configurations. 

\begin{table*}[h]
\centering
\scriptsize
\setlength{\tabcolsep}{3pt}
\begin{tabular}{lrcrrrrr}
\toprule
Submission & $L$ & Observed interval & $R_k$ & $S_{\rm avail}$ & $A_{\rm avail}$ & $S_6$ & $A_6$ \\
\midrule
AEGIS & 7 & [0.000, 100.000] & 100.000 & +0.4030 & 1.2652 & +0.3948 & 1.3522 \\
$^{\dagger}$COMETKiwi22 & 7 & [0.342, 0.907] & 0.565 & +0.0044 & 0.0064 & +0.0045 & 0.0067 \\
Cohere CAT+ ensemble & 7 & [18.333, 100.000] & 81.667 & +2.2575 & 3.9098 & +2.3355 & 4.0877 \\
FACET\_869543 & 7 & [-42.000, 0.000] & 42.000 & +0.7989 & 1.7933 & +0.7698 & 1.7619 \\
FACET\_869546 & 7 & [-42.000, 0.000] & 42.000 & +0.8178 & 1.8556 & +0.7867 & 1.8280 \\
$^{\dagger}$Gemini-3.6-Flash& 6 & [0.000, 100.000] & 100.000 & +1.8793 & 3.9061 & +1.8793 & 3.9061 \\
$^{\dagger}$Gemma 4 & 6 & [0.000, 100.000] & 100.000 & +4.5768 & 6.8781 & +4.5768 & 6.8781 \\
$^{\dagger}$Gemma 4 RB & 6 & [0.000, 100.000] & 100.000 & +4.5173 & 6.8458 & +4.5173 & 6.8458 \\
Lexicala (Majority-Vote) & 1 & [25.770, 100.000] & 74.230 & +8.5108 & 10.8195 & -- & -- \\
Lexicala (STAPLE) & 1 & [23.890, 100.000] & 76.110 & +9.2165 & 11.9635 & -- & -- \\
MQM-LLM & 7 & [6.769, 98.928] & 92.159 & +3.2272 & 4.8277 & +3.3506 & 4.9895 \\
MQM-LLM CA & 7 & [14.518, 102.219] & 87.700 & +3.3401 & 5.1159 & +3.4152 & 5.2599 \\
$^{\dagger}$Qwen 3.6& 6 & [0.000, 100.000] & 100.000 & +4.0614 & 7.3392 & +4.0614 & 7.3392 \\
$^{\dagger}$Qwen 3.6 RB & 6 & [0.000, 100.000] & 100.000 & +3.9874 & 7.2333 & +3.9874 & 7.2333 \\
Vertical\_870257 & 7 & [0.122, 0.965] & 0.843 & +0.0038 & 0.0147 & +0.0027 & 0.0148 \\
Vertical\_870554 & 7 & [0.000, 1.000] & 1.000 & +0.0052 & 0.0251 & +0.0022 & 0.0239 \\
bytepop\_pro10 & 7 & [0.000, 100.000] & 100.000 & +2.5534 & 3.9723 & +2.3525 & 3.7023 \\
bytepop\_pro9 & 7 & [0.000, 100.000] & 100.000 & +2.4174 & 3.8296 & +2.1938 & 3.5359 \\
cuni-v14-regression & 1 & [43.393, 82.535] & 39.141 & +3.2708 & 4.5328 & -- & -- \\
fluency2-gemini35-esa & 7 & [54.895, 100.000] & 45.105 & -0.0081 & 5.9563 & -0.0996 & 5.7234 \\
gemba-poly & 7 & [0.000, 100.000] & 100.000 & +0.3099 & 1.3081 & +0.3421 & 1.3192 \\
$^{\dagger}$xCOMET XL RB & 6 & [14.248, 100.000] & 85.752 & +2.1348 & 3.2522 & +2.1348 & 3.2522 \\
$^{\dagger}$xCOMET XL RF & 6 & [14.248, 100.000] & 85.752 & +2.1348 & 3.2522 & +2.1348 & 3.2522 \\
$^{\dagger}$xCOMET XXL RB & 6 & [18.329, 100.000] & 81.671 & +4.6963 & 5.7238 & +4.6963 & 5.7238 \\
$^{\dagger}$xCOMET XXL RF & 6 & [18.329, 100.000] & 81.671 & +4.6963 & 5.7238 & +4.6963 & 5.7238 \\
\bottomrule
\end{tabular}
\caption{Observed score scales and macro means in native units. $L$ is the number of returned target languages. Absolute means are means of pairwise absolute differences, not absolute values of signed means. $^{\dagger}$ denotes a baseline; unmarked systems are participant submissions.}
\label{tab:native-scales}
\end{table*}

\end{document}